\documentclass{article}

\usepackage{microtype}
\usepackage{graphicx}
\usepackage{subcaption}
\usepackage{booktabs}
\usepackage{hyperref}
\usepackage{xcolor}

\usepackage[accepted]{icml2026}
\makeatletter
\renewcommand{\Notice@String}{\textit{Accepted to the ICML 2026 AI4Research Workshop on AI as a Tool for Mathematics, Computer Science, and Machine Learning.}}
\makeatother

\usepackage{amsmath}
\usepackage{amssymb}
\usepackage{mathtools}
\usepackage{amsthm}
\usepackage[capitalize,noabbrev]{cleveref}
\newcommand{\cmark}{\ensuremath{\checkmark}}
\newcommand{\xmark}{\ensuremath{\times}}

\theoremstyle{definition}

\icmltitlerunning{Negative Knowledge as Failure-aware Shared Memory for AutoResearch}

\begin{document}

\twocolumn[
  \icmltitle{Negative Knowledge as Failure-aware Shared Memory for AutoResearch}

  \begin{icmlauthorlist}
    \icmlauthor{Hanchun Wang}{damtp}
  \end{icmlauthorlist}

  \icmlaffiliation{damtp}{Department of Applied Mathematics and Theoretical Physics, University of Cambridge, Cambridge, UK}

  \icmlcorrespondingauthor{Hanchun Wang}{hw660@cam.ac.uk}

  \icmlkeywords{AI-assisted research, scientific workflows,
    research graphs, human-AI collaboration, reproducibility,
    negative results}

  \vskip 0.3in
]

\printAffiliationsAndNotice{}
\begin{abstract}
AI-assisted research systems generate many failed attempts, but those
failures rarely become a durable, shared knowledge asset. We propose a \emph{negative knowledge memory layer}: a curator agent converts each failed attempt into a bounded, typed record in a shared bank, and a downstream research agent explicitly adopts or rejects those records before proposing its next experiment. 
We evaluate this layer in two settings: same-task retry on ScienceAgentBench and cross-task scientific research on two nonlinear math-physics PDE problems. The negative knowledge layer outperforms vanilla AutoResearch baselines while using fewer tokens; agents with the negative knowledge bank solve new tasks that all baselines fail to solve in PDE systems research. We also show that the previous negative knowledge bank can transfer and enhance AutoResearch on different PDE problems. These results suggest that structured negative knowledge is \emph{a knowledge asset that should be explicitly maintained} in broader AI-engaged scientific research beyond a memory-compression or debugging aid, alongside positive findings, as a collective infrastructure for scientific memory.  Code is available at \url{https://github.com/hch-wang/Negative_Knowledge}.
\end{abstract}

\section{Introduction}
\label{sec:intro}
AutoResearch systems aim to automate or semi-automate parts of the
scientific research process---literature retrieval, hypothesis
generation, experiment planning and execution, result analysis,
review, and writing
\citep{lu2024aiscientist,schmidgall2025agentlab,gottweis2025aicoscientist,chen2024scienceagentbench,gu2024blade,boiko2023coscientist,bran2024chemcrow}.

\begin{figure*}[h]
    \centering
    \includegraphics[width=\linewidth]{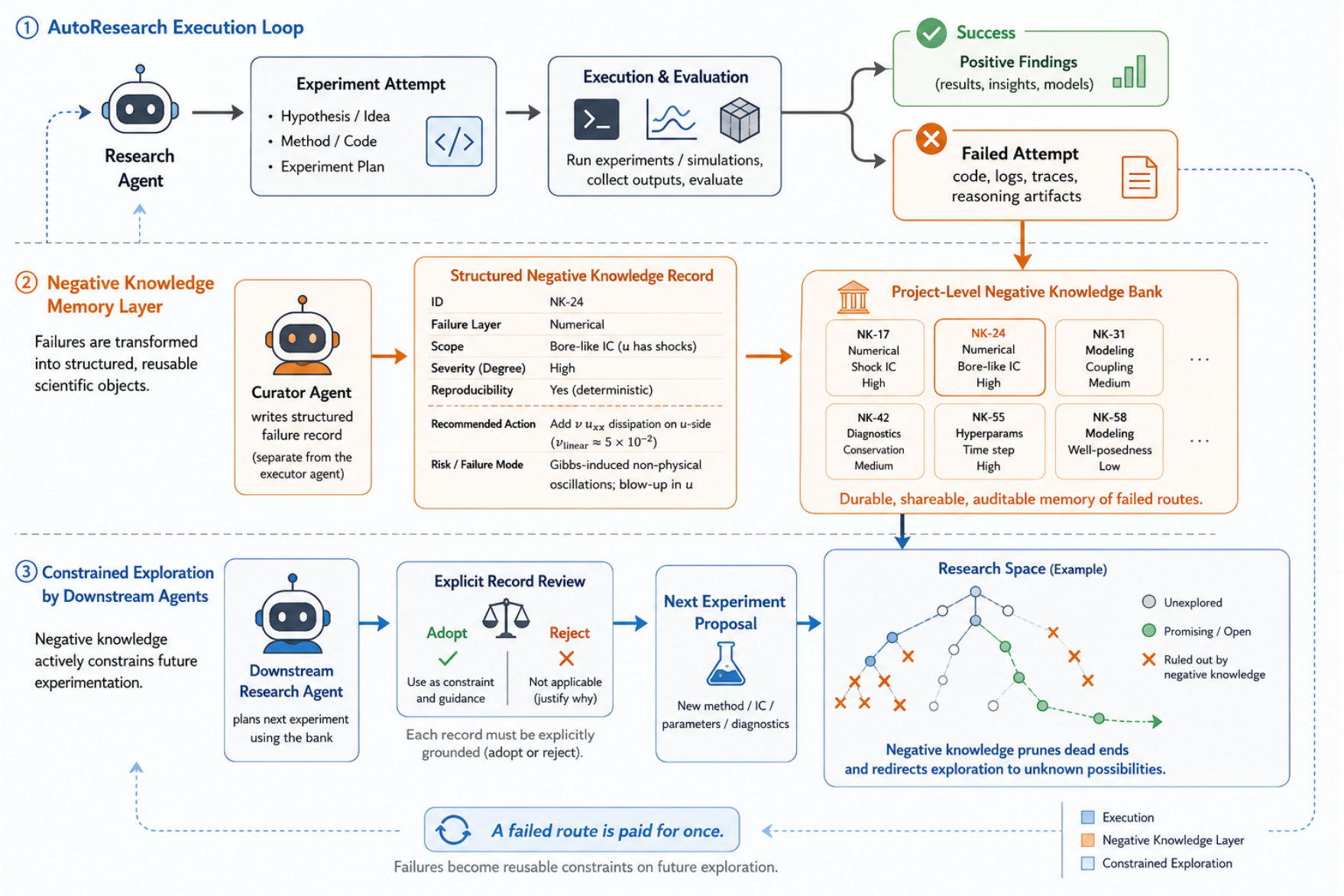}
    \caption{Overview of the proposed negative knowledge memory layer in a multi-agent AutoResearch workflow. A research agent proposes and executes experiments, producing both positive findings and failed attempts. Instead of discarding failures as transient failure signals, a separate curator agent converts the artifacts of a failed attempt (e.g., code, logs, traces, reasoning outputs) into a bounded and typed negative knowledge record in a shared project-level bank. Before proposing a new experiment, downstream research agents must explicitly inspect the bank and ground their planning decisions by adopting or rejecting relevant records. In this way, failures become reusable constraints on future exploration: previously explored dead ends can be avoided, while experimental search is redirected toward unexplored regions of the research space. The figure emphasizes the central claim of this work: negative knowledge functions not merely as memory augmentation, but as shared epistemic infrastructure for autonomous scientific research.
}
    \label{fig:fig1}
\end{figure*}

However, the dominant orientation of AutoResearch remains
success-seeking. Existing workflows primarily ask how prior successful
knowledge can be retrieved, recombined, and extended into new
successful results; failures appear as local debugging signals but
rarely become durable research objects. Recent agent frameworks let
failures inform the next in-session attempt via verbal reflection or
self-debug \citep{shinn2023reflexion,madaan2023selfrefine,chen2023selfdebug},
but the failed route itself is not preserved as a cross-agent,
cross-task object. Using negative or counterexample signals to constrain
search appears in safe and counterexample-guided
reinforcement learning and probabilistic verification
\citep{alshiekh2018safe,ji2023counterexample,ji2025robust}---but there the
negative signal is consumed within a single learning process. This leaves out a second form
of scientific memory: \emph{negative knowledge}---failed hypotheses,
non-working methods, abandoned experimental routes, rejected
interpretations, boundary conditions, and unresolved anomalies.

The focus on positive knowledge alone reflects a historical publication bias, extensively discussed in the
philosophy of science
\citep{popper1959logic,kuhn1962structure,lakatos1970falsification,
rosenthal1979filedrawer,ioannidis2005false}. For human scholars,
emphasising failed routes carries low individual reward; failed work
therefore survives primarily as informal tacit knowledge circulated
within laboratories and expert communities
\citep{polanyi1966tacit,collins2010tacit}. The prevailing scientific
paradigm offers no structural reward for human scholars to spend
resources producing, preserving, or sharing negative knowledge. The
cost is collective inefficiency: a failed route has little value as
an individual success claim, yet substantial value for the research
community at large.

In multi-agent AutoResearch, an agent does not need human-style publication
credit, which changes the underlying game-theoretic relation. An agent can therefore be
assigned a different epistemic role---a system-level subject that spends real resources to generate and maintain negative knowledge as a shared asset. For human scholars, sustained low-reward labor of this kind is difficult. Even for LLM-based systems the obstacle is reward design itself: rewarding both positive and negative outcomes dilutes the reward signal, while rewarding only positive findings may lead to false progress. Splitting these roles across multiple agents is therefore a sound system design.

We therefore propose a minimal framework for maintaining negative
knowledge: agents create, maintain, and inspect negative knowledge
inline with their AutoResearch workflow, and humans can intervene to
revise it at any point. This paper studies a small but operational
first step---a project-level AutoResearch setting in which a few AI
agents and human researchers jointly maintain reviewable negative
knowledge. The broader vision is that the scientific community itself
can be understood as the largest AutoResearch system: once AI agents
remove researcher productivity as a socially scarce resource,
negative knowledge can become maintainable and shareable public
epistemic infrastructure.

This paper makes three contributions. First, we define an
architecture-agnostic negative knowledge memory layer for
human-audited agent-assisted research. Second, we introduce a bounded
failure schema that turns failed routes into structured research
objects and inter-agent communication resources. Third, we report a
pilot evaluation on a scientific coding benchmark and a PDE
numerical-methods case study, measuring whether structured negative knowledge predicts failure repairability, reduces token cost in multi-round retry, and supports multi-agent sharing.

\section{Method: Negative Knowledge Memory Layer}
\label{sec:workflow}

We introduce a \emph{negative knowledge memory layer} that turns each failed attempt into a reusable record rather than a transient log entry. Unlike prior agent-memory architectures that accumulate successful skills, episodic interactions, or long-term context \citep{wang2024voyager,park2023generativeagents,packer2023memgpt}, this layer is explicitly designed for failure. Our method combines a structured negative knowledge record with a multi-agent workflow \citep{wu2023autogen}: a \emph{curator agent} writes records from finished-attempt artifacts into a shared \emph{bank}, and a \emph{research agent} reads the bank before proposing the next experiment.

The schema has three design properties.
(1) \emph{Attempt-level}: each record describes one finished attempt.
(2) \emph{Bounded}: records follow prompt-level length and structure
limits, so an entire bank can be passed to an agent inline.
(3) \emph{Typed}: the failure-classification fields take their values
from closed vocabularies, so the curator must commit to an explicit
failure class rather than describe the failure in free prose.
These properties make records portable, auditable, and reusable across
attempts. Each record contains bounded fields for the failure layer,
scope, degree, recommended action, and risk; \Cref{appendix:schema}
provides a detailed example.

The \textbf{curator} is a separate agent from the one
that ran the attempt; once that attempt finishes, the curator combines
the artifacts it left behind (e.g., the executed code, execution output,
or reasoning notes) into one schema-conforming record. This separation reduces self-assessment bias: the curator sees frozen artifacts rather than the failed agent's own near-miss narrative.

The \textbf{research agent} then reads from the knowledge bank before launching its attempt under an explicit grounding requirement: each record must
be marked as adopted or rejected, with a justification tying the
decision to the specific failure mode the record describes. This
turns documented failures into active constraints on the next
attempt. The research agent's output is the next attempt
itself, whose artifacts feed the curator pass that follows and add
a new record to the bank, closing the workflow loop.

The central goal is for a scientific project to pay for a failed route
only once, then redirect later attempts toward routes not yet ruled out.
We evaluate this design in same-task retry and cross-task transfer
settings.

\section{Evaluation: Negative-Knowledge Retry}\label{sec:evaluation}
This section evaluates the negative knowledge layer in a controlled
same-task retry setting on a scientific-coding benchmark. We test whether
a structured record of a failed attempt helps a fresh agent repair the
same task without access to the original conversation. We compare
negative-knowledge retry with direct retry and self-debug baselines,
measuring pass rate and the size of the memory object handed to the
next attempt.

\paragraph{Experimental setup.} All experiments in this section use Claude Sonnet~4.6 \citep{anthropic2026sonnet46}
on the deterministic-evaluation subset of ScienceAgentBench
\citep{chen2024scienceagentbench}, where outcomes are verified by
programmatic evaluators. We compare five conditions: \texttt{Base} (first attempt),
\texttt{Retry} (fresh agent, no memory; identical to \texttt{Base} up to
resampling),
\texttt{Self-debug}~\citep{chen2023selfdebug} (three rounds of raw
failure feedback: each later attempt sees the previous code and execution output), \texttt{Negative knowledge retry} (one structured record), and \texttt{Deep negative knowledge retry} (one record distilled
from three failed rounds). We report \emph{pass rate} (\% tasks passing the deterministic evaluator) and \emph{memory size}, the tokens of the memory object shown to the next attempt. Full benchmark and baseline details are in \cref{appendix:scienceagentbench}.
\begin{table}[h]
\centering
\caption{Performance on deterministic-eval ScienceAgentBench. Negative knowledge retry uses one failed round, self-debug uses three rounds, and deep negative knowledge retry distills three rounds into one record.}
\label{tab:section3-main}
\resizebox{0.99\columnwidth}{!}{
\begin{tabular}{lcc}
\toprule
Method & Pass rate (\%) & Memory (tokens)\\
\midrule
Base & 31.6 & -- \\
Retry & 31.6 & -- \\
Negative knowledge retry & 36.8 & 296 \;($-73.3\%$) \\
Self-debug & 44.7 & 1,109 \;(baseline) \\
Deep negative knowledge retry & \textbf{47.4} & 795 \;($-28.3\%$) \\
\bottomrule
\end{tabular}
}
\end{table}
\paragraph{Main result.} \Cref{tab:section3-main} shows three things. First, plain retry does not improve performance, but adding one structured negative-knowledge record does:  \texttt{Base} reaches a pass rate of $31.6\%$, and \texttt{Retry} remains at $31.6\%$; by contrast, \texttt{Negative knowledge retry} raises the pass rate to $36.8\%$. The gain comes from an explicit record of what failed and how to change course, not from retrying itself. Second, distilling three failed rounds into one \texttt{Deep negative knowledge} record ($47.4\%$) outperforms raw three-round self-debug ($44.7\%$), with the gap concentrated on the hard subset where raw self-debug has already failed; this suggests structured negative knowledge can carry guidance that raw feedback alone does not. Third, negative-knowledge records achieve these gains while using $28.3\%$ and $73.3\%$ fewer tokens than self-debug, so the layer preserves useful repair information in a more compact form.

\section{Case Study: Negative Knowledge in a Mathematical Physics Research Loop}
\label{sec:case-study}
This section tests whether negative knowledge can serve as reusable shared failure memory for a research team by transferring from related exploratory tasks to similar new tasks and reshaping agents' experimental choices. We evaluate structured negative knowledge in an open-ended research workflow on the coupled Burgers-swept-KdV (BKdV) system introduced by \citet{holm2025compound}. BKdV is a useful testbed for two reasons. First, it was introduced recently, has no standard computational treatment in the literature, and the phenomena we study remain open questions, leaving no established answer for an agent to retrieve. Second, it combines several challenging numerical PDE features, including shocks, solitons, dispersion, stiffness, and nonlinear coupling. The BKdV system couples a Burgers-like current $u$ to a KdV-like wave $v$; the reduction $u = v^2/2 $ collapses to a Gardner equation; the three limits (Burgers, KdV, Gardner) are known while the nonlinear phenomena are still open questions:
\begin{equation}
\begin{split}
          u_t + 3 u u_x &= -\partial_x\!\left(3 v^2 + \gamma v_{xx}\right), \\
  v_t + 6 v v_x + \gamma v_{xxx} &= -\partial_x(u v).
\end{split}
\end{equation}
To study the solution behaviour of the system, we use two stages; all case-study sub-agents run on Claude Sonnet~4.5.
\textbf{Stage~1} builds a shared knowledge bank from multi-round stress tests on the coupled
system and its reduced limits. \textbf{Stage~2} asks whether agents can use that bank to make better
experimental choices on three new coupled-system research questions. Full task
specifications, bank inventory, AutoResearch protocol, evaluation details, and trace excerpts are in
\cref{appendix:bkdv-setup}.

\subsection{Stage 1: Building the knowledge bank}
\label{sec:pde-stage1}
In Stage 1, to build the knowledge bank, we ran AutoResearch-style three-round stress
tests on Burgers, KdV, Gardner, shallow-water, and coupled BKdV settings. The
resulting bank contains 58 records: 15 positive and 43 negative, including
7 depth-3 path-closure records. These records are a project-local memory of what the agents
had already learned, including which routes worked, which routes failed, and
which failures should constrain later experiments. Two examples illustrate the
kind of reusable knowledge stored in the bank: a negative record (BKdV-S6)
establishing the failure boundary of a pre-validated numerical stack on
bore-like initial conditions, and a positive Gardner record showing that
a method validated on one reduced limit transfers cleanly to another.
Full entries are in \cref{appendix:bkdv-bank}.

\subsection{Stage 2: Research on BKdV nonlinear phenomena}
\label{sec:pde-stage2}
Stage~2 tests whether the knowledge bank transfers to three new mathematical
physics sub-tasks in the coupled BKdV system: \textbf{Test-A} soliton stability
near the $m=0$ manifold, \textbf{Test-B} Gaussian wave-packet decomposition into
a soliton train, and \textbf{Test-C} KdV-soliton interaction with a Burgers bore.
We compare four conditions applied on the AutoResearch pipeline: \textbf{Base} (no bank), \textbf{Base+Pos}
(positive records only), \textbf{Base+Neg} (negative bank only), and
\textbf{Base+Pos+Neg} (full bank). Each cell is allowed up to three AutoResearch rounds and is judged by deterministic physics-aware checks; detailed definitions and settings are in
\cref{appendix:bkdv-setup}.
\begin{table}
\centering
\caption{Transfer of the knowledge bank to three BKdV sub-tasks.}
\label{tab:bkdv-headline}
\small
\begin{tabular}{lcccc}
\toprule
Condition & Test-A& Test-B& Test-C& Success \\
\midrule
Base          & \xmark{} & \xmark{} & \xmark{} & 0/3 \\
Base+Pos      & \xmark{} & \xmark{} & \cmark{} & 1/3 \\
Base+Neg      & \cmark{} & \cmark{} & \cmark{} & \textbf{3/3} \\
Base+Pos+Neg  & \cmark{} & \cmark{} & \cmark{} & \textbf{3/3} \\
\bottomrule
\end{tabular}
\end{table}

\subsection{Trace analysis: how negative knowledge changes a proposal}
\label{sec:tc-mechanism}
To inspect the mechanism behind the table, we examine the Test-C/Base+Neg
trace. After the baseline run on the bore-soliton interaction exhibits
bore-driven instability, the agent consults the relevant negative-knowledge
record, follows its prescription by changing only one component of the
previous proposal, and explicitly rejects two alternative routes the bank
also rules out. The cell then succeeds in two rounds. This shows the
operational role of negative knowledge: not merely a warning, but a reusable
constraint that narrows the next experimental proposal.

The weaker Base+Pos result reflects a feature particular to
numerical-simulation research: successful methods do not always transfer
between related problems. A positive Gardner record can reasonably motivate
a validated setup for BKdV, but in Test-B this transfer encounters a
stronger coupled-system failure mode. Negative records are more reusable
because they warn where the transfer breaks; adding them to the positive
bank (Base+Pos+Neg) restores the performance. Verbatim trace excerpts for
both patterns are in \cref{appendix:agent-traces}.

\subsection{Cross-system transfer to Burgers-NLS}
\label{sec:bnls-transfer}

We also test whether the BKdV knowledge bank can help on a substantially
different coupled system, Burgers-NLS (BNLS) \citep{dombret2023collisions}. In a three-round AutoResearch
setting, agents without any bank fail all four BNLS tasks, while agents given
only the BKdV bank solve two of them. This suggests that the bank is not only
task-local memory: some of its numerical failures, diagnostics, and method
warnings transfer to a different nonlinear PDE system and become a useful
collective asset for the research team. Details of the BNLS setup
are in \cref{appendix:bnls-transfer}.

\section{Conclusion}
\label{sec:conclusion}
We introduced a negative knowledge memory layer: a curator agent writes each failed attempt as a bounded, typed record into a shared bank, and a downstream research agent must adopt or reject those records before its next experiment. On ScienceAgentBench, structured records improve retry pass rate at lower token cost than raw self-debug; on two coupled PDE systems, the bank enables cross-task transfer that no-bank baselines fail. Within the limits of small sample sizes and the models studied, this is a first step toward maintaining negative knowledge as shared epistemic infrastructure for AI-engaged scientific research.

\bibliography{references}
\bibliographystyle{icml2026}

\newpage
\appendix
\onecolumn

\section{Negative-Knowledge Record Schema}
\label{appendix:schema}

A negative-knowledge (NK) record is a compact description of a failed,
partial, or misleading route. The record has three roles at once: it
summarises the attempted route, classifies the failure in a closed
taxonomy, and gives the next agent a concrete alternative or boundary.
The base record contains
\texttt{task\_id}, \texttt{attempted\_route}, \texttt{observation},
\texttt{failure}, \texttt{rationale}, and
\texttt{recommended\_alternative}. The nested \texttt{failure} object is
the typed part of the schema. Failure classification is judged field by
field: layer, scope, degree, action, and risk.

\begin{table}[h]
\centering
\caption{Controlled vocabulary for the negative-knowledge failure fields.}
\label{tab:nk-schema-fields}
\small
\begin{tabular}{p{0.18\textwidth}p{0.74\textwidth}}
\toprule
Field & Allowed values and meaning \\
\midrule
\texttt{layer} &
\texttt{implementation\_failure},
\texttt{communication\_failure},
\texttt{method\_failure}. The PDE curators additionally permit
\texttt{hypothesis\_failure} and \texttt{measurement\_failure} when the
failure is scientific rather than software-local. \\
\texttt{scope} &
\texttt{local\_failure},
\texttt{regime\_bound\_failure},
\texttt{general\_failure}. \\
\texttt{degree} &
\texttt{contradicted},
\texttt{partial},
\texttt{inconclusive},
\texttt{unstable},
\texttt{artifact\_driven},
\texttt{overclaimed}. \\
\texttt{action} &
\texttt{retry},
\texttt{change\_method},
\texttt{narrow\_claim},
\texttt{abandon\_route}. In JSON records this is stored as
\texttt{recommended\_action}. \\
\texttt{risk} &
\texttt{low\_risk\_omission},
\texttt{medium\_risk\_drift},
\texttt{high\_risk\_false\_progress}. \\
\bottomrule
\end{tabular}
\end{table}

The schema intentionally separates \emph{classification} from
\emph{prescription}. The classification fields tell a downstream agent
what kind of failure was observed and how strongly it should constrain
future attempts. The prescription fields
\texttt{rationale} and \texttt{recommended\_alternative} state why the
route failed and what the next agent should try instead. This prevents
the bank from becoming only a veto list: a negative record can rule out a
route and still name a constructive replacement.

Depth-$N$ records add cross-round synthesis fields:
\texttt{depth}, \texttt{rounds\_summary},
\texttt{ruled\_out\_routes}, and
\texttt{synthesised\_diagnosis}. These fields are used when the curator
reads multiple failed rounds from the same task. In that case,
\texttt{attempted\_route} and \texttt{observation} move into
\texttt{rounds\_summary}, while the top-level diagnosis records the shared
mechanism across all failed routes.

\section{ScienceAgentBench Retry Details}
\label{appendix:scienceagentbench}
\subsection{Benchmark and Baselines}
ScienceAgentBench~\citep{chen2024scienceagentbench} is a benchmark
for evaluating language agents on data-driven scientific discovery
tasks drawn from real scientific workflows. It contains 102 tasks
extracted from peer-reviewed scientific publications and evaluates
whether an agent can produce an executable solution for each task. We
use its deterministic-evaluation subset of 38 tasks to ensure all
reported outcomes are verified by programmatic evaluators.

We compare five conditions, which differ only in whether and how
failure information is shown to a subsequent attempt. Three are
baselines and two use the negative knowledge layer:
\begin{itemize}\itemsep0pt
  \item \texttt{Base}: the first attempt on each task, without any
        additional failure memory.
  \item \texttt{Retry}: a fresh agent invocation on the same task,
        without access to the previous conversation, code, outputs, or
        negative-knowledge memory.
  \item \texttt{Self-debug}~\citep{chen2023selfdebug,chen2024scienceagentbench}:
        later attempts receive raw feedback from prior failed attempts,
        including previous code and execution or evaluation output, for
        up to three rounds.
  \item \texttt{Negative knowledge retry}: the agent receives one
        structured negative-knowledge record written by the curator from
        the failed \texttt{Base} attempt.
  \item \texttt{Deep negative knowledge retry}: the agent receives one
        structured negative-knowledge record distilled by the curator
        from three failed rounds.
\end{itemize}

We report \emph{pass rate}, the percentage of tasks that pass the
deterministic evaluator, and \emph{memory size}, the size of the additional
memory object shown to the next agent, measured with
the \texttt{cl100k\_base} tokenizer. Memory size captures what a
workflow must store and re-send in order to reuse a failure; it is not the
end-to-end cost of a condition. From the released dispatch logs, the median
end-to-end token usage per task (all rounds, plus curation, plus the final
attempt) is ${\approx}56$k for three-round self-debug, ${\approx}53$k for
depth-1 negative-knowledge retry (first attempt, a curator pass of
${\approx}19$k, and one retry), and ${\approx}101$k for deep
negative-knowledge retry, which consumes the same three failed rounds plus a
deep curator pass (${\approx}27$k) and one further attempt. Curation is
therefore a real cost; the compact record amortises it when a failure is
stored, shared, or consulted more than once. \texttt{Base} and \texttt{Retry}
have no additional memory object.

All ScienceAgentBench sub-agent prompts use the same fixed template.
The template contains four blocks: the task instruction copied from the
benchmark, a data block with the folder tree and dataset preview, an
output-path specification, and a memory block. Only the memory block
changes across conditions. It is empty for \texttt{Base} and
\texttt{Retry}; contains raw prior-failure feedback for
\texttt{Self-debug}; contains one bounded negative-knowledge record for
\texttt{Negative knowledge retry}; and contains one distilled
three-round record for \texttt{Deep negative knowledge retry}.
Tasks where \texttt{Self-debug} fails all three rounds form the
\emph{hard subset} referenced in \cref{sec:evaluation}.

\subsection{Example: Task 072 in ScienceAgentBench}
We use task~072 as an example of how deep negative knowledge can change
the repair strategy. This task involves EEG signal mapping from subject
01 to subject 03. The
three-round \texttt{Self-debug} baseline does not pass this task, while
\texttt{Deep negative knowledge retry} passes the evaluator by replacing repeated neural-network retries with a closed-form
linear-regression solution.

\begin{table}[tb]
\centering
\caption{Per-round trace for task~072. The first three rounds follow
the \texttt{Self-debug} condition (round~1 with no memory, rounds~2--3
with raw feedback from prior rounds) and repeatedly attempt PyTorch
U-Net variants, each of which times out on CPU. The final round uses
only the distilled negative-knowledge record. The research agent follows
the record's recommendation and implements a closed-form per-channel
least-squares solution, which passes the evaluator.}
\label{tab:appendix-task072}
\footnotesize
\begin{tabular}{cp{0.30\textwidth}p{0.36\textwidth}p{0.12\textwidth}}
\toprule
Round & Memory shown to the agent & Attempted strategy & Outcome \\
\midrule
1 &
No prior failure memory &
PyTorch U-Net with batch size 64, 30 epochs, and Adam optimizer with
learning rate $10^{-3}$ ($\approx 7{,}800$ steps) &
Timeout \\
2 &
Code and error output from round 1 &
Larger batch size of 512, 8 epochs, and OneCycle learning-rate schedule
($\approx 260$ steps) &
Timeout \\
3 &
Code and error output from round 2 &
Smaller 3-stage U-Net, reduced from 4 stages, with a
\texttt{ConvTranspose1d} decoder &
Timeout\\
4 &
Only the distilled negative-knowledge record &
Per-channel least-squares regression:
\texttt{np.linalg.lstsq} applied separately to each of the 17 channels &
\textbf{Pass} ($0.763$) \\
\bottomrule
\end{tabular}
\end{table}

The curator record diagnosed the shared failure mode across the first
three rounds as a wall-clock bottleneck from CPU-bound tensor
computation. All three failed attempts used variants of an 8-layer
one-dimensional convolutional U-Net on $16{,}540 \times 17 \times 200$
float32 inputs, so the curator treated them as the same failed route
rather than as independent implementation choices. The record therefore
recommended a different computational regime: closed-form per-channel
linear regression. The downstream research agent adopted this
recommendation and passed the task.

\section{BKdV Case Study Details}
\label{appendix:bkdv-setup}

\begin{figure}[h]
\centering
\includegraphics[width=.8\textwidth]{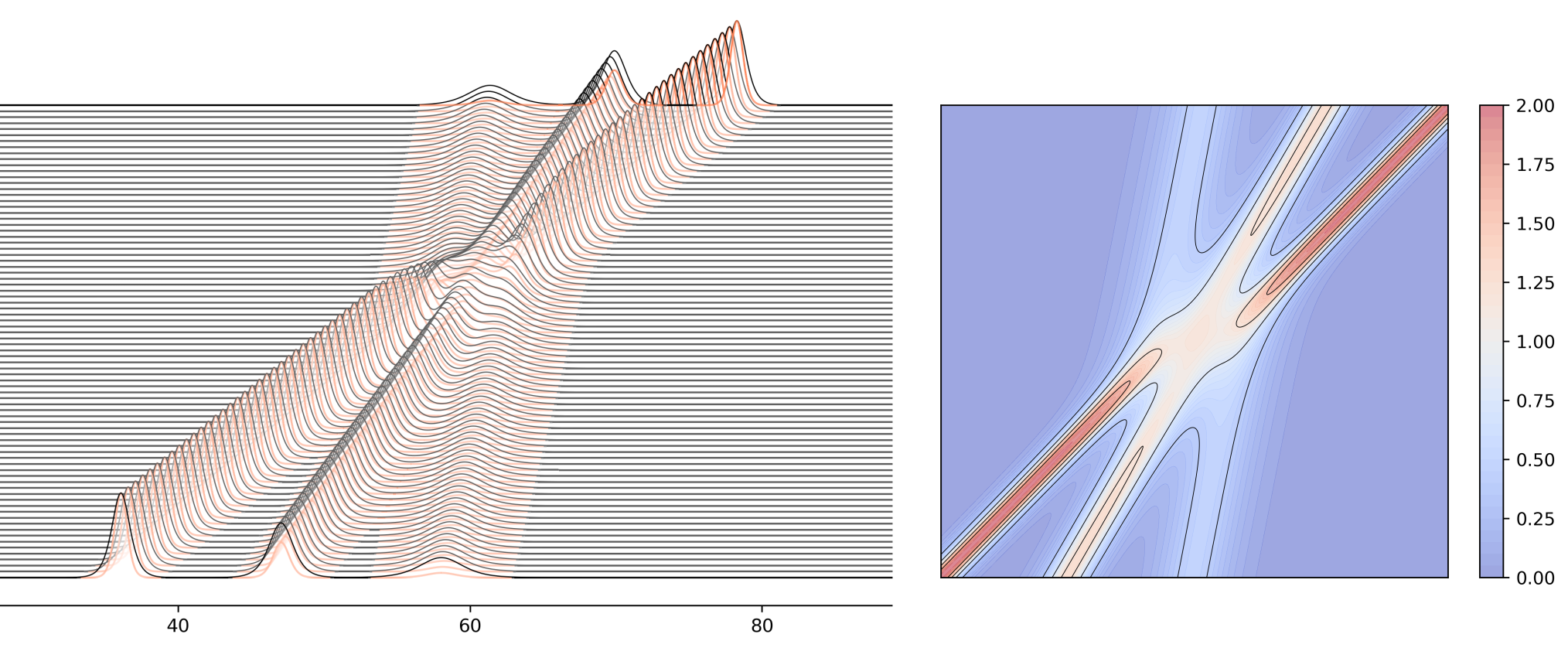}
\caption{\textbf{Compound solitons on the $u = v^2/2$ manifold (context for
Test-A).} Propagation and collision of three compound solitons of the BKdV
system on the invariant manifold $u = v^2/2$. \emph{Left:} waterfall plot (time
increasing upward) of the Burgers mean-flow velocity $u$ (red) and the KdV wave
field $v$ (black); the locked $u$--$v$ pulses travel at near-constant speed and
survive their mutual collisions. \emph{Right:} contour plot of $v(x,t)$ tracing
the three soliton trajectories and their interaction. Figure reproduced from
\citet{holm2025compound}; it shows the stable solutions whose off-manifold
robustness Test-A probes, not an agent output.}
\label{fig:bkdv-overview}
\end{figure}

\begin{figure}[h]
\centering
\includegraphics[width=\textwidth]{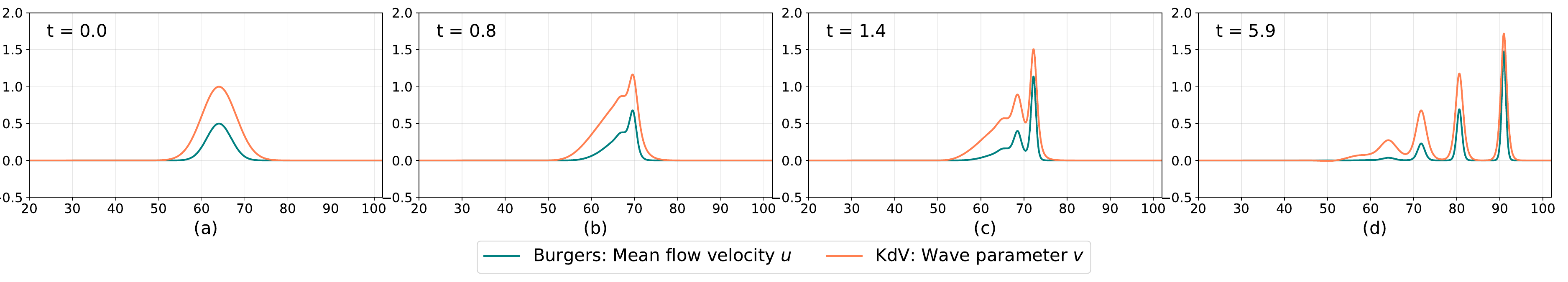}
\caption{\textbf{Soliton fission (basis of Test-B).} An initial Gaussian wave
packet in the BKdV system steepens and decomposes into a rank-ordered train of
compound solitons. Panels (a)--(d) show the Burgers mean-flow velocity $u$
(green) and the KdV wave field $v$ (orange) at $t = 0,\,0.8,\,1.4,\,5.9$. Figure
reproduced from \citet{holm2025compound}; it depicts the target phenomenology
for Test-B (Gaussian wave-packet decomposition), not an agent output.}
\label{fig:bkdv-fission}
\end{figure}

\begin{figure}[h]
\centering
\includegraphics[width=\textwidth]{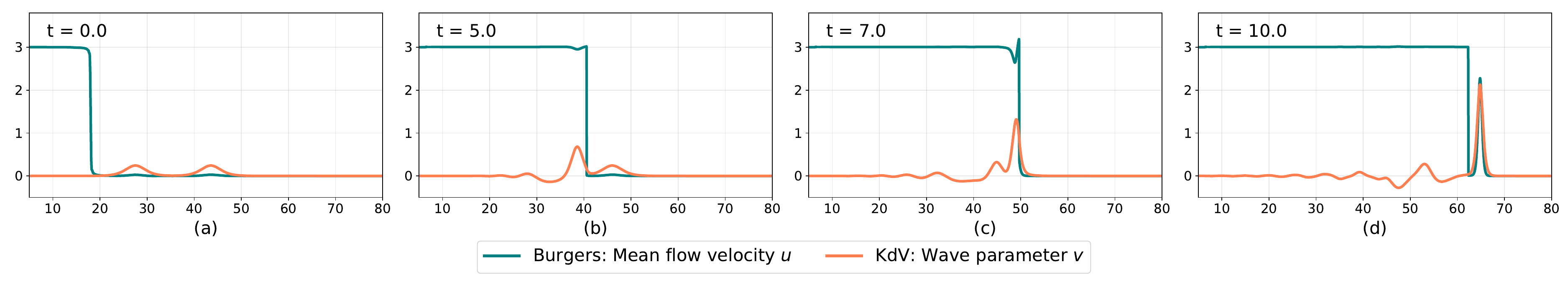}
\caption{\textbf{Bore--soliton fusion (basis of Test-C).} A Burgers bore (the
$u \approx 3$ front, green) overtakes two slower, weak KdV solitons (orange) and
sweeps them together, fusing them into a single compound soliton that rides at
the bore front; snapshots at $t = 0,\,5,\,7,\,10$. Figure reproduced from
\citet{holm2025compound}; it depicts the target bore--soliton interaction for
Test-C (KdV-soliton interaction with a Burgers bore), not an agent output.}
\label{fig:bkdv-bore}
\end{figure}

\paragraph{The BKdV system and its phenomenology.}
Our case study is set on the compound Burgers--swept-KdV (BKdV) system of
\citet{holm2025compound}, which couples a Burgers-type mean-flow velocity
$u(x,t)$ to a KdV-type wave field $v(x,t)$,
\begin{equation}
\label{eq:bkdv-appendix}
\begin{split}
  u_t + 3 u u_x &= -\partial_x\!\left(3 v^2 + \gamma v_{xx}\right), \\
  v_t + 6 v v_x + \gamma v_{xxx} &= -\partial_x(u v),
\end{split}
\end{equation}
where $\gamma$ sets the strength of the KdV dispersion and the cross term
$-\partial_x(u v)$ provides the two-way coupling between the two fields. The
system is a demanding numerical testbed because it interpolates between three
classical integrable limits---Burgers (shock formation), KdV (solitons), and,
on the invariant manifold $u = v^2/2$, the Gardner equation---yet its fully
coupled, off-manifold behaviour has no standard computational treatment and
remains an open research question. On the invariant manifold the system admits
stable \emph{compound solitons}---localised waves in which the Burgers mean flow
and the KdV wave lock together---that propagate and collide cleanly
(\cref{fig:bkdv-overview}); whether they remain stable when the initial state is
pushed slightly off the manifold is the subject of Test-A. The fully coupled
dynamics produce two further phenomena that our Stage-2 tasks ask agents to study
numerically. First, a smooth wave
packet undergoes \emph{soliton fission}, steepening and breaking up into a
rank-ordered train of compound solitons (\cref{fig:bkdv-fission}, the basis of
Test-B). Second, a Burgers \emph{bore} interacts with KdV solitons by
refracting, reflecting, or fusing them, so that an overtaking bore can sweep
several weak solitons into a single compound soliton at its front
(\cref{fig:bkdv-bore}, the basis of Test-C). \Cref{fig:bkdv-overview,fig:bkdv-fission,fig:bkdv-bore}
reproduce this reference phenomenology from \citet{holm2025compound}: they show
the target behaviour a successful agent must recover numerically, and are not
agent outputs.

\subsection{Research Graph Protocol}
\label{appendix:research-graph}

Both Stage~1 stress-test programs and Stage~2 cells in the BKdV case study
run sub-agents under a shared \emph{Research Graph} protocol---an
AutoResearch loop in the spirit of reason-and-act agent pipelines
\citep{yao2022react}, where in each round an agent proposes an experiment
from the current research question and context, runs the corresponding
numerical computation, records the resulting finding, and decides whether
to revise the method, narrow the question, or stop.

\paragraph{Node types.} A sub-agent maintains
\texttt{research\_state.jsonl}, an append-only event log, with four node
types:
\begin{itemize}\itemsep0pt
  \item \textbf{Question (Q)}: a research question to answer.
  \item \textbf{Experiment (E)}: a concrete numerical experiment---a specific
        (IC, method, parameters, $T$) tuple to be executed.
  \item \textbf{Finding (F)}: the observed outcome of an experiment
        (numerical diagnostics, interpretation, and a self-assessment).
  \item \textbf{Decision (D)}: a research-direction choice
        (\texttt{retry}, \texttt{change\_method}, \texttt{narrow\_claim},
        \texttt{abandon\_route}, or \texttt{stop\_useful}) derived from
        one or more Findings.
\end{itemize}

\paragraph{Round budget.} One \emph{round} is one E node plus one
execution of \texttt{candidate.py} plus one F node. Each cell is allowed
up to three rounds. Bug-fix re-runs (typos, undefined variables) that
test the same E design do not count as a new round.

\paragraph{Bank consultation requirement.} For bank-aware conditions
(\texttt{Base+Pos}, \texttt{Base+Neg}, \texttt{Base+Pos+Neg}), every E
node must populate three fields:
\begin{itemize}\itemsep0pt
  \item \texttt{cites\_bank}: list of bank entry IDs the proposal leverages
        (e.g.\ \texttt{["BKdV-S6-deep"]}).
  \item \texttt{rejects\_bank}: list of bank entry IDs the proposal
        explicitly avoids.
  \item \texttt{bank\_use\_rationale}: one-sentence justification of how
        the cited and rejected entries shaped the proposal.
\end{itemize}
These three fields make the bank's role auditable at proposal time; the
trace excerpts in \cref{appendix:agent-traces} are drawn directly from
\texttt{bank\_use\_rationale} fields written during the runs.

\subsection{Stage 1: Knowledge Bank Construction}
\label{appendix:bkdv-bank}

\paragraph{Stage~1 stress tests.}
\textbf{Burgers shock} ($u_t + u u_x = 0$, $u_0 = -\sin(\pi x)$,
periodic on $[-1,1]$, $N_x{=}200$): A1 forced forward-Euler +
central FD at $T{=}0.5$; A2 any stable scheme at $T{=}0.1$
(pre-shock); A3 any stable scheme at $T{=}10$ (long-time
contamination). \textbf{KdV soliton}
($v_t + 6 v v_x + v_{xxx} = 0$, $v_0 = 2\,\mathrm{sech}^2(x+5)$,
periodic on $[-15,15]$, $N_x{=}256$, $T{=}2$): A4 forced explicit
RK4 + central FD for $v_{xxx}$; A5 forced Fourier spectral with
\emph{no} dealiasing; A6 forced IC amplitude $0.1$ (small-amplitude
regime). \textbf{Shallow water dam-break} ($h_t + (hu)_x = 0$,
$(hu)_t + (hu^2 + g h^2/2)_x = 0$, $g{=}1$, $h_L{=}2$, $h_R{=}1$,
$T{=}0.4$): A7 forced forward-Euler + central FD; A8 forced
Lax-Friedrichs; A9 dry-bed initial condition ($h_R{=}0$); A10
forced HLL. \textbf{Gardner}
($v_t + 6 v v_x + \tfrac{3}{2} v^2 v_x + v_{xxx} = 0$, $T{=}2$):
G1 forced explicit RK4 (IC amp $1.5$); G2 IMEX-CN spectral with
2/3 dealiasing (IC amp $1.5$); G3 IMEX-CN spectral with no
dealiasing (IC amp $1.5$); G4 IMEX-CN spectral with IC amp $3.0$
(amplitude-CFL test). \textbf{Coupled BKdV} (system as defined in
\cref{sec:case-study}): S1 numerical-method survey at amplitudes
$1$--$3$ over $T{=}10$; S2 conservation-law audit (divergence-form
trivialities $\int u\,dx, \int v\,dx$ vs.\ physical non-conservations
$\int u v\,dx$ and energy candidates); S3 IC-family dependence
(broadband seeds at amplitude $\geq 0.8$ hit a high-k cascade);
S4 resolution sensitivity at $N_x{=}256$, defining a hyperviscosity
safe envelope $\nu_h \lesssim 10^{-20}$ for smooth ICs;
S5 $m{=}0$ manifold (non-)invariance; S6 bore-IC u-viscosity
necessity (pseudospectral + 2/3 dealias + RK4 is quantitatively wrong
for bore-like $u$, producing Gibbs-driven excursions; usable
$\nu_\text{linear} \approx 5\times 10^{-2}$, 13 orders above the S4
envelope); S7 Gardner-stable $\not\Rightarrow$ BKdV-stable ($-62.8\%$
$v_{\max}$ decay).

\paragraph{Example positive and negative records.}
\begin{itemize}
  \item \textbf{Numerical-method failure.} BKdV-S6 establishes that the
        pre-validated stack (pseudospectral + 2/3 dealias + RK4) is
        \emph{quantitatively wrong} for bore-like $u$ initial conditions:
        $u$ develops Gibbs-driven non-physical excursions, and a usable
        threshold is $\nu_\text{linear} \approx 5\times 10^{-2}$ (or
        $\nu_h \approx 10^{-9}$ for $k^8$ hyperviscosity), \emph{13 orders of
        magnitude} above the BKdV-S4 smooth-soliton ``safe envelope''.
  \item \textbf{Possible feasible method.} A Gardner record states that
        IMEX-Crank--Nicolson spectral with 2/3 dealiasing, first validated on
        KdV at amplitude $2.0$ and $\Delta t=5\times 10^{-4}$, transfers
        cleanly to Gardner at moderate amplitude
        ($v=1.5\,\mathrm{sech}^2$, $T=2$): mass is conserved, amplitude is
        preserved, and the solution remains a single peak.
\end{itemize}

\subsection{Stage 2: Sub-task Setup and Evaluation}
\label{appendix:eval-v2}

\paragraph{Stage~2 sub-task specifications.}
\textbf{Test-A soliton stability:}
$u_0 = v_0^2/2 + 0.2\, v_0$, $v_0 = 2\,\mathrm{sech}^2(x+5)$, $T{=}8$.
\textbf{Test-B Soliton Fission:}
$v_0 = 4\exp(-(x+5)^2/2.25)$, $u_0 = 0$, $T{=}6$.
\textbf{Test-C bore-soliton interaction:}
$u_0 = 1.5\cdot(1 - \tanh(x/0.5))/2$ on $[-15,15]$,
$v_0 = 1.5\,\mathrm{sech}^2(x+8)$, $T{=}8$. Each cell has a budget of
up to three rounds with a \texttt{finding\_record} carried across
rounds within a cell.

\paragraph{Evaluation criteria.} A cell counts as useful only if its
integration completes without NaN and total mass drift stays below
$8\%$. The remaining task-specific gates are:
\begin{itemize}\itemsep0pt
  \item \textbf{Test-A} (soliton stability):
        $\max(|u|,|v|) < 15$,
        $v_{\max}(T)/v_{\max}(0) \geq 0.25$, and a single dominant peak
        ($n_\text{peaks}({\geq}0.4) = 1$ or top-to-second peak ratio
        $> 1.5$). The amplitude floor is calibrated to BKdV-S7's
        quantified $-62.8\%$ off-manifold decay; the single-peak gate
        rejects chaotic fragmentation.
  \item \textbf{Test-B} (soliton-train fission):
        $\max(|u|,|v|) < 15$, $v_{\max}(T) \geq 0.8$, and at least two
        dominant peaks ($n_\text{peaks}({\geq}0.8) \geq 2$).
  \item \textbf{Test-C} (bore-soliton interaction):
        $\max(|u|,|v|) < 10$ with the additional constraint $u_{\max} < 5$
        (the bore must remain bounded; IC has $u_L = 1.5$),
        $v_{\max}(T) \geq 0.5$, and at least one dominant peak
        ($n_\text{peaks}({\geq}0.5) \geq 1$).
\end{itemize}

\subsection{Agent Trace Catalogue}
\label{appendix:agent-traces}

We extract verbatim text from \texttt{research\_state.jsonl} files of
selected cells to illustrate how bank-aware agents use the bank. Each
excerpt is the agent's own field text, written during the run.

\paragraph{Excerpt~1: Test-C/PosNeg E2 \texttt{bank\_use\_rationale}.}
The cell passes at round~2 by adopting BKdV-S6's prescriptive $\nu_\text{lin}$
while explicitly rejecting BKdV-S4's hyperviscosity envelope:

\begin{quote}\small
``POSITIVE bank guidance: BKdV-S6 r=3 establishes $\nu_\text{lin} \in
\{10^{-2}, 5\cdot 10^{-2}\}$ both pass; $\nu = 5\cdot 10^{-2}$ yields
$\text{TV}_\text{final}{\sim}9.6$ and $u_\text{min,final} \geq 0$ (best).
NEGATIVE bank guidance: BKdV-S6 deep synthesis explicitly warns that
BKdV-S4's safe envelope ($\nu_h \sim 10^{-22}$ to $10^{-20}$ for $k^8$
hyperviscosity) is 13 orders of magnitude too weak for bore-like IC---we
REJECT transferring that envelope. We also reject the would-be-tempting
alternative of just shrinking dt (E1 already used dt$=10^{-4}$ which is
below post-dealias CFL---dt alone cannot stop the bore cascade per
BKdV-S6 r=2 finding that $\varepsilon=10^{-4}$ viscosity was ineffective).''
\end{quote}

This is the prescription-and-warning pattern in action: the agent reads
\emph{both} which level to use AND which would-be-tempting shortcut to skip.

\paragraph{Excerpt~2: Test-A/PosOnly F2 rationale (physics-honest fail).}
Under the expanded bank including BKdV-S7, PosOnly no longer applies the
hyperviscosity ``trick'' that artificially preserves $v_{\max}$ above
threshold; instead it diagnoses the failure as physical:

\begin{quote}\small
``Stack ran cleanly to $T=8$ with mass conserved to machine precision and
$|$fields$|$ bounded. However the soliton is destroyed:
$v_{\max}(T)=0.917 < 1.0$ ($0.46\times$ of initial $2.0$), and $v$ contains
5--6 peaks above $0.7$. This matches the BKdV-S7 r2 physical breakdown of
the m{=}0 manifold under the $-\partial_x(uv)$ coupling---and our IC is more
perturbed ($m_0 = +0.2 v_0 = {\sim}0.4 \text{sech}^2$, not~0). The result
\emph{may} still satisfy the [\ldots]''
\end{quote}

The agent now \emph{cites} the BKdV-S7 mechanism instead of attempting a
numerical workaround. This is a research-character outcome:
bank explanation of why something fails physically discourages numerical
sleight-of-hand.

\paragraph{Excerpt~3: Test-C/NegOnly E2 \texttt{bank\_use\_rationale}.}
NegOnly demonstrates that a sufficiently prescriptive negative bank carries
its own actionable content (it does not need positive entries to recommend
a method):

\begin{quote}\small
``Single-component upgrade vs E1: add $\nu \cdot u_{xx}$ on $u$ only
(treated explicitly inside RK4). BKdV-S6 deep synthesis: $\nu=5\cdot 10^{-2}$
empirically passes ($\text{TV}_\text{final} \sim 9.6$,
$u_\text{min,final} \geq 0$) on the EXACT same bore IC used here, while the
alternative hyperviscosity $\nu_h$ must approach the explicit-RK4 stability
ceiling ${\sim}10^{-9}$ (10+ orders above BKdV-S4 `safe envelope' $10^{-22}$).
REJECT BKdV-S4 ladder: $\nu_h \leq 10^{-12}$ is empirically insufficient
($\text{TV}>115$, $u_\text{max}>3$)---would waste an iteration. REJECT
MUSCL/Godunov on $u u_x$: a 2+ component swap from E1 (changes both
discretisation AND adds limiter) violates progressive-complexity [\ldots]''
\end{quote}

This is what we mean by ``prescriptive negative'': the entry both \emph{rules
out a tempting wrong direction} (transferring the BKdV-S4 envelope) and
\emph{names a working level} ($\nu = 5\cdot 10^{-2}$). Without the named
level the bank would still rule out BKdV-S4's $\nu_h \sim 10^{-22}$, but
the agent would have no anchor for what level \emph{does} work.

\subsection{Burgers-NLS Cross-System Transfer}
\label{appendix:bnls-transfer}

The Burgers-NLS (BNLS) appendix tests whether a bank built on BKdV can
transfer to a related but mechanistically different coupled PDE. The
BNLS system couples a Burgers current $u$ to a focusing NLS field in
Madelung form $\Psi = \sqrt{N}\,e^{i\phi}$, giving an $(u, N, \phi)$
triple with a compound-soliton manifold $M_\mathrm{cs} = \{u = N\,\phi_x\}$.
In Madelung variables the density $N$ is advected by the flow and the phase
obeys
\begin{equation*}
  \phi_t + u\,\phi_x = -\tfrac{1}{2}\phi_x^2 - \frac{(\sqrt{N})_{xx}}{2\sqrt{N}} + F'(N),
\end{equation*}
so dispersion enters through the \emph{quantum-pressure} term
$-(\sqrt{N})_{xx}/2\sqrt{N}$, $F'(N)$ is a focusing nonlinearity, and the
current $u$ is Burgers-swept as before. 

\emph{Similar to BKdV but not the same:} both couple a shock-forming Burgers bore to a dispersive nonlinear wave
on a compound-soliton manifold---so the Burgers-side numerics carry over---but
the wave sectors differ. BKdV's $v$ is a real KdV field (third-derivative
dispersion $v_{xxx}$, rank-ordered soliton trains), whereas BNLS's wave is the
complex NLS field above (quantum-pressure dispersion and a focusing cubic,
giving modulational instability and envelope solitons). This shared skeleton
with a different wave mechanism is what makes BNLS a stringent transfer target.

The study uses four BNLS tasks and four memory conditions: no bank,
BKdV-only bank, NLS-specific bank, and the combined NLS+BKdV bank;
all BNLS sub-agents run on Claude Sonnet~4.6. The
NLS bank contains 21 entries curated from eight BNLS stress tests; the
BKdV bank reuses a 30-entry snapshot of the BKdV study's bank
(10 positive and 20 negative entries).

\paragraph{Sub-task specifications.} Each task runs on a periodic
domain $[-15, 15]$ with $N_x = 256$ and saves snapshots of
$(u, N, \phi)$; PASS is decided by a deterministic phenomenon check
on the final snapshot.
\textbf{Test-A NLS-soliton stability on $M_\mathrm{cs}$:}
bright soliton $N(x,0) = A^2\,\mathrm{sech}^2(A(x+5))$ with $A = 1.5$
and $u = N\,\phi_x$ exactly (so $m_0 = 0$), $T = 8$. Useful iff mass
drift $< 5\%$, fields bounded, and the final $N$ contains a single
peak with amplitude $\geq 0.5\times$ the initial $N$-max.
\textbf{Test-B Gaussian-packet modulational instability:}
Gaussian density $N_0 = 2.0\exp(-(x+5)^2/2.25)$ on $M_\mathrm{cs}$,
$T = 6$. Useful iff mass drift $< 5\%$, fields bounded, and the final
$N$ has $\geq 2$ well-separated peaks with amplitude $\geq 1.0$
(soliton-train emission).
\textbf{Test-C bore-soliton interaction:}
Burgers bore $u_0 = (1 - \tanh(x/0.5))/2$ plus a bright soliton
$N_0 = \mathrm{sech}^2(x+8)$ off $M_\mathrm{cs}$, $T = 8$. Useful iff
$u$ stays bounded ($|u|_\mathrm{max} < 5$) and the final $N$ contains
a peak with amplitude $\geq 0.3$ (soliton survives the interaction).
\textbf{Test-D compound-soliton attractor relaxation (research-grade):}
bright soliton plus an off-manifold perturbation
$u_0 = N\,\phi_x + \varepsilon\cos(2\pi x/L)$ with
$\varepsilon \in \{0.05, 0.1, 0.2, 0.4\}$, $T = 12$. Useful iff the
integration is numerically stable (mass drift $< 5\%$, fields bounded);
the research deliverable is a characterisation of $\|m\|_2(t)$
relaxation (decay / plateau / growth), where $m = u - N\,\phi_x$.

\begin{table}[h]
\centering
\caption{BNLS cross-system transfer. The domain-matched NLS bank solves
three of four tasks; the BKdV-only bank transfers partially but also creates
negative transfer on one task.}
\label{tab:bnls-transfer}
\small
\begin{tabular}{lccccc}
\toprule
Condition & Test-A & Test-B & Test-C & Test-D & Success \\
\midrule
No bank       & \xmark{} & \xmark{} & \xmark{} & \xmark{} & 0/4 \\
BKdV only     & \cmark{} & \xmark{} & \xmark{} & \cmark{} & 2/4 \\
NLS only      & \cmark{} & \cmark{} & \cmark{} & \xmark{} & \textbf{3/4} \\
NLS + BKdV    & \cmark{} & \cmark{} & \cmark{} & \xmark{} & \textbf{3/4} \\
\bottomrule
\end{tabular}
\end{table}

The headline result is that domain-matched negative knowledge is the most
reliable transfer signal: the NLS bank lifts the no-bank baseline from
0/4 to 3/4. Adding the BKdV bank to the NLS bank gives no additional task
success in this run, suggesting that cross-domain numerical knowledge is
useful only when the receiving agent can distinguish transferable
mechanisms from mismatched ones.

The BKdV-only condition is informative because it both helps and misleads.
It solves two tasks, showing that some Burgers-side numerical lessons
transfer. On Test-B, however, BKdV-only knowledge produces negative transfer:
the agent follows a method family that is validated for BKdV-like dynamics
but anti-diffusive under the BNLS variational sign convention. 
\end{document}